\documentclass[11pt]{article}
\usepackage{acl}

\usepackage{times}
\usepackage{latexsym}
\usepackage[T2A,T1]{fontenc}
\usepackage[utf8]{inputenc}
\usepackage[russian,english]{babel}
\newcommand{\cyrtext}[1]{\foreignlanguage{russian}{#1}}

\usepackage{microtype}
\usepackage{inconsolata}
\usepackage{graphicx}
\usepackage{booktabs}
\usepackage{multirow}
\usepackage{url}
\usepackage{longtable}
\usepackage{float}
\newenvironment{exquote}
  {\par\addvspace{1pt}\begingroup\small
   \leftskip=1.2em \rightskip=0pt \parindent=0pt \noindent\ignorespaces}
  {\par\endgroup\addvspace{1pt}}
\usepackage{tikz}
\usetikzlibrary{arrows.meta,positioning,calc,fit,backgrounds}

\title{Do Rule-Based Tools Help LLMs Translate Endangered\\
Uralic Languages in the Prompt or over MCP?}

\author{
  \textbf{Iaroslav Chelombitko\textsuperscript{1,2}},
  \textbf{Mika H{\"a}m{\"a}l{\"a}inen\textsuperscript{1}}
\\
  \textsuperscript{1}Metropolia University of Applied Sciences, Helsinki, Finland \\
  \textsuperscript{2}Neapolis University Pafos, Paphos, Cyprus
}

\begin{document}
\maketitle

\begin{abstract}
Komi-Zyrian, Moksha, Erzya and Udmurt have mature rule-based resources, among
them finite-state analysers, constraint grammars and bilingual dictionaries,
which language models do not use. We ask whether giving a model access to
these resources improves machine translation from the four languages into
English, and whether it matters how the access is provided: as analyses
computed in advance and written into the prompt, or as tools that the model
calls itself through the Model Context Protocol (MCP). We evaluate manually:
an expert linguist read every English output of the strongest model under
three conditions, 816 sentences in all, accepted or rejected each one, and
named the main error in every rejected translation. With the analyses in the prompt, the
number of accepted translations rises from 13 to 43 out of 272; with the tools
attached it rises to 55. The gain comes from the bilingual dictionary. For
Udmurt, whose dictionary has no English side, the prompt-based condition brings
no gain at all, while for the other three languages an English gloss reaches
the prompt for 74 to 80\% of the source words. Tool calling is also much
harder to get working than prompt injection. Of 31 open-weight models, 14 never
call a tool, 11 of them because they cannot produce a usable call, and half of
the calls that are made name a language the toolkit does not support. We
release the data set, every model output and all tool-call logs.
\end{abstract}

\section{Introduction}

Komi-Zyrian, Moksha, Erzya and Udmurt are poorly supported by current
language models \citep{purason2025finnougric}. At the same time they are well
described. Decades of finite-state and constraint-grammar work
\citep{rueter2021komi,rueter2020mordvinic} have produced morphological
analysers, generators and bilingual dictionaries for each of them. These
resources are maintained in the GiellaLT infrastructure
\citep{moshagen2023giellalt} and are available through the UralicNLP Python
library \citep{hamalainen2019uralicnlp}. UralicMCP
\citep{hamalainen2025uralicmcp} proposes to bring the two together by exposing
the resources to a language model as tools that the model can call.

So far the proposal has been demonstrated rather than evaluated, and it
leaves a design question open. There are two ways to bring linguistic
knowledge to a model.
The analyses can be computed in advance and written into the prompt, which
requires nothing of the model beyond reading them. Alternatively, the tools
can be attached to the model, which then decides what to look up; the Model
Context Protocol \citep{mcp2024} makes this easy to set up. The second option
is the more attractive one, since it scales to inputs that nobody has
pre-processed and since it is how agentic systems are being built today.

We compare the two options on a 68-sentence parallel test set in the four
languages, with 31 open-weight instruction-tuned models of up to 27B
parameters and seven experimental conditions. We decided early on to rely
on manual evaluation. An expert linguist read every English output of the
strongest model, Qwen3.8-27B, under three conditions, accepted or rejected
each output, and assigned an error category to every rejected one. The reason
is that the only reference translation we have is a free literary one, and
against such a reference an automatic metric mostly measures how closely a
system happens to match one translator's wording. We still report chrF++ and
BLEU in Appendix~\ref{sec:scores} so that readers can compare with other
work, but none of our claims rests on them.

Writing the analyses into the prompt raises the number of accepted
translations from 13 to 43 out of 272. The gain is confined to the languages
whose dictionary has an English side; for Udmurt, where no English gloss
reaches the prompt, there is none. Giving the same model the tools instead
raises the count further, to 55, but this comparison is only possible for the
one model in the roster that calls the tools on every sentence and still
answers in prose. Across the roster the picture is different. 14 of the 31
models never emit a tool call, 11 of them because they cannot produce one that
a parser reads, and even the models that do call send an unsupported language
code in half of their calls. Translation into the four languages, finally,
fails under every condition, with 23\% of the outputs in the wrong language
and 13\% stuck in a repetition loop. What is missing in that direction is a
model of the target language itself, and the rule-based resources cannot
supply it.

The rest of the paper is organised as follows. Section~\ref{sec:related} reviews related work, Section~\ref{sec:data}
describes the data, Section~\ref{sec:pipeline} the conditions and models,
Section~\ref{sec:error-analysis} the manual evaluation,
Section~\ref{sec:roster} the tool-calling behaviour of the full roster, and
Section~\ref{sec:discussion} concludes.

\section{Related work}
\label{sec:related}

The Finno-Ugric languages are severely underrepresented in language model
development \citep{purason2025finnougric}, and the gap is visible already at
the level of the tokeniser \citep{chelombitko2024tokenizers}.
\citet{hamalainen2021endangered} argues that calling these languages
``low-resource'' is misleading, since they lack text corpora rather than
linguistic description. Two kinds of response have been proposed.
The first changes the model: \citet{purason2025finnougric} continue
pretraining and instruction-tune models for V\~oro, Livonian and Komi. The
second leaves the model unchanged and changes what it receives at inference
time. \citet{hamalainen2024dag} take this route by putting the dictionary
translations of candidate lemmas into a ChatGPT prompt in order to
disambiguate Erzya and Skolt Sami sentences, reaching 41\% and 50\% accuracy,
and UralicMCP \citep{hamalainen2025uralicmcp} goes one step further by letting
the model fetch the material itself.

The study closest to ours is \citet{tereschenko2025gpt}, who evaluate GPT
models on translation between Finnish and the same four languages, compare
reasoning and non-reasoning architectures, and report a 16-point difference in
refusal rates. That work uses frontier APIs and no linguistic tools. Earlier
work on machine translation of Uralic languages includes rule-based systems
\citep{johnson2017north,rueter2021prerequisites,pirinen2022building} and
neural systems \citep{yankovskaya2023machine,dale2022first,alnajjar2023bootstrapping}.

Token fertility as a measure of how well a tokeniser represents a language
follows \citet{rust2021tokenizer}. For Uralic languages,
\citet{chelombitko2024tokenizers} quantify the imbalance and show that it can
be remedied, reducing the compression ratio for Finnish from 3.41 under
LLaMA-2 to 1.18 with a monolingual BPE tokeniser; \citet{xu2026tokenization}
find that morphologically better-aligned segmentation improves downstream
tagging across six Uralic languages; and \citet{chelombitko2025samponlp}
derive per-language vocabulary-size recommendations from the lexicon.

\section{Data}
\label{sec:data}

We use the 68-sentence parallel set of literary prose introduced by
\citet{alnajjar-etal-2023-sentiment}, which covers English, Finnish,
Komi-Zyrian, Moksha, Erzya and Udmurt. The set is small, but to our knowledge
it is the only one that is parallel across all four endangered languages with
a single English source and checked translations. One English text was
translated independently into each language, so the English side is identical
across the five Uralic pairs. Each sentence carries a sentiment polarity label
(35 negative, 33 positive); we do not use these labels in this paper, and
the Limitations explain why.\footnote{The set with its English side, together
with all model outputs and logs, is released at
\url{https://huggingface.co/datasets/Aragoner/UralicMCPBench}. The Uralic and
Finnish sides are those of \citet{alnajjar-etal-2023-sentiment}, archived at
\url{https://doi.org/10.5281/zenodo.7866456} without the English translation.}

The morphological analysers cover the set well. They return at least one
reading for 91.1\% of the Udmurt word tokens and for 95.4\% to 98.5\% of the
tokens in the other four languages, so the results below cannot be explained
by the tools failing to parse the input.

\section{Experimental setup}
\label{sec:pipeline}

\definecolor{pSrc}{RGB}{248,215,218}
\definecolor{pDat}{RGB}{198,225,246}
\definecolor{pPro}{RGB}{217,215,240}
\definecolor{pMod}{RGB}{251,223,190}
\definecolor{pTool}{RGB}{238,241,194}
\definecolor{pOut}{RGB}{202,226,202}

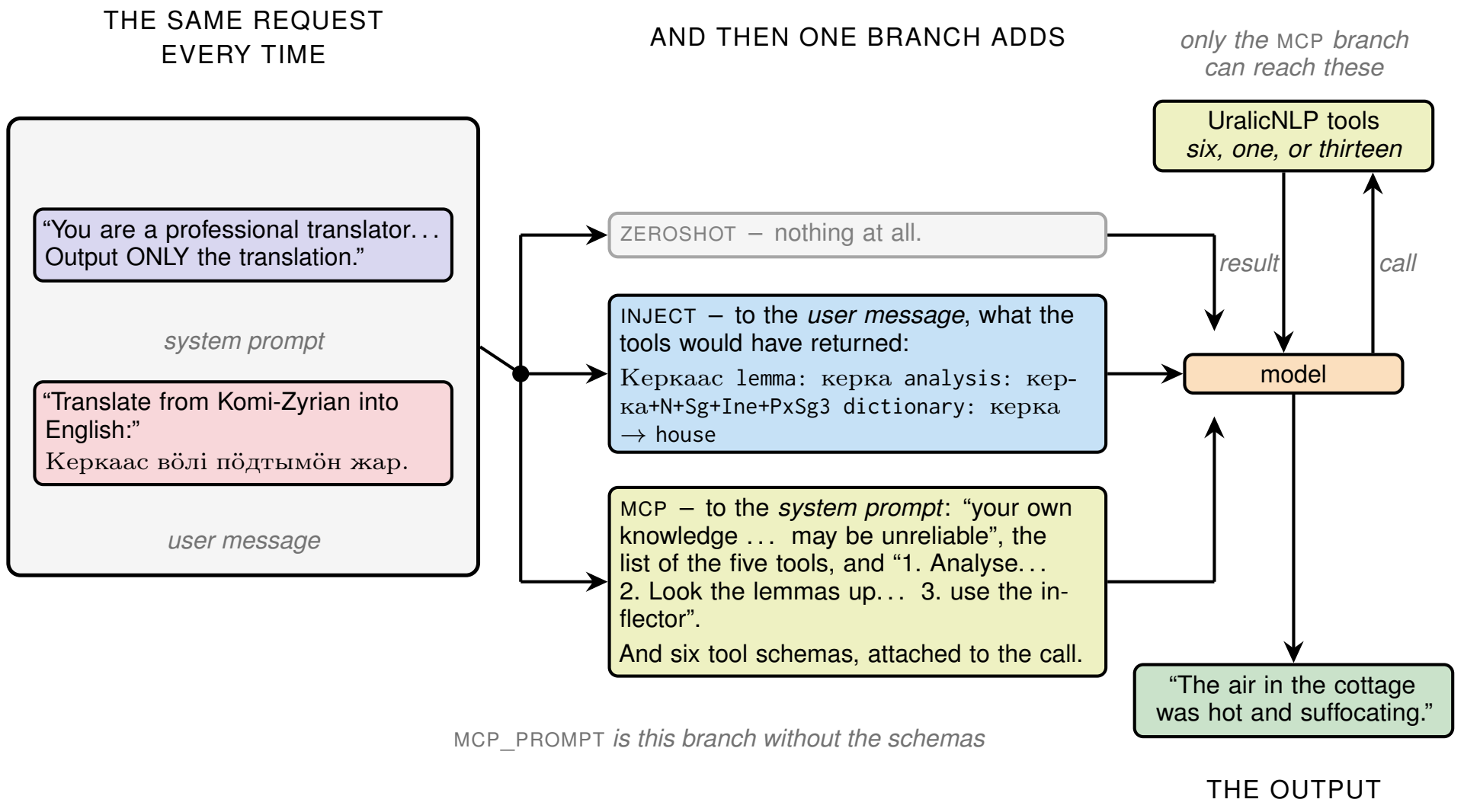
\begin{figure*}[t]
\centering
\scalebox{0.86}{
\begin{tikzpicture}[
  x=1mm, y=1mm, font=\sffamily,
  bx/.style   = {draw, line width=0.9pt, rounded corners=2.5pt, align=left,
                 font=\sffamily\scriptsize, inner sep=3pt, text width=40mm},
  br/.style   = {bx, text width=48mm},
  ctr/.style  = {draw, line width=0.9pt, rounded corners=2.5pt, align=center,
                 font=\sffamily\scriptsize, inner sep=3pt},
  hd/.style   = {font=\sffamily\footnotesize\scshape, align=center},
  bdg/.style  = {font=\sffamily\scriptsize\itshape, text=black!55, align=center},
  arr/.style  = {-{Stealth[length=2.2mm,width=2mm]}, line width=0.9pt,
                 rounded corners=2.5pt}
]

\node[hd] at (22,42) {the same request\\every time};
\node[bx, fill=pPro] (b1) at (22,21)
     {``You are a professional translator\ldots{} Output ONLY the
       translation.''};
\node[bdg, anchor=north] at (22,13.5) {system prompt};
\node[bx, fill=pSrc] (b2) at (22,2)
     {``Translate from Komi-Zyrian into English:''\\[1pt]
      \cyrtext{Керкаас вӧлі пӧдтымӧн жар.}};
\node[bdg, anchor=north] at (22,-7) {user message};

\node[hd] at (84,42) {and then one branch adds};

\node[br, fill=black!4, draw=black!35, text=black!50] (c1) at (84,22)
     {\textsc{zeroshot} \,--\, nothing at all.};

\node[br, fill=pDat] (c2) at (84,8)
     {\textsc{inject} \,--\, to the \emph{user message}, what the tools would
      have returned:\\[2pt]
      {\ttfamily\scriptsize \cyrtext{Керкаас}\, lemma: \cyrtext{керка}\, analysis:
       \cyrtext{керка}+N+Sg+Ine+PxSg3\, dictionary: \cyrtext{керка} $\to$ house}};

\node[br, fill=pTool] (c3) at (84,-13)
     {\textsc{mcp} \,--\, to the \emph{system prompt}: ``your own knowledge
      \ldots{} may be unreliable'', the list of the five tools, and
      ``1.~Analyse\ldots{} 2.~Look the lemmas up\ldots{} 3.~use the
      inflector''.\\[2pt] And six tool schemas, attached to the call.};
\node[bdg, anchor=north] at (70,-26.5) {\textsc{mcp\_prompt} is this branch
      without the schemas};

\begin{scope}[on background layer]
  \node[draw, line width=1.1pt, rounded corners=3.5pt, fill=black!4,
        fit=(b1)(b2), inner xsep=2.5mm, inner ysep=9mm] (req) {};
\end{scope}
\draw[line width=0.9pt] (req.east) -- (50,8);
\draw[line width=0.9pt] (50,22) -- (50,-13);
\draw[arr] (50,22) -- (c1.west);
\draw[arr] (50,8)  -- (c2.west);
\draw[arr] (50,-13) -- (c3.west);
\fill (50,8) circle (0.85mm);

\node[ctr, fill=pMod, minimum width=22mm] (mod) at (128,8) {model};
\draw[line width=0.9pt] (c1.east) -- (120,22);
\draw[line width=0.9pt] (c3.east) -- (120,-13);
\draw[arr] (120,22) -- (120,12.3);
\draw[arr] (120,-13) -- (120,3.7);
\draw[arr] (c2.east) -- (mod.west);

\node[ctr, fill=pTool, text width=26mm] (too) at (128,32)
     {UralicNLP tools\\{\scriptsize\itshape six, one, or thirteen}};
\draw[arr] ([xshift=8mm]mod.north) -- node[bdg, right, inner sep=1.2pt] {call}
           ([xshift=8mm]too.south);
\draw[arr] ([xshift=-1mm]too.south) -- node[bdg, left, inner sep=1.2pt] {result}
           ([xshift=-1mm]mod.north);
\node[bdg, anchor=north, text width=34mm] at (128,44)
     {only the \textsc{mcp} branch can reach these};

\node[ctr, fill=pOut, text width=30mm] (ans) at (128,-25)
     {``The air in the cottage was hot and suffocating.''};
\draw[arr] (mod.south) -- (ans.north);
\node[hd] at (128,-34) {the output};

\end{tikzpicture}}
\caption{How conditions are built.}
\label{fig:conditions}
\end{figure*}

Every condition sends the same request to the same model: a system prompt
that asks for a translation and nothing else, and a user message with the
source sentence. The conditions differ only in what is added to this request.
Figure~\ref{fig:conditions} shows the shared request and the additions made by
the \textsc{inject} and \textsc{mcp} branches, and Table~\ref{tab:conditions}
lists all seven conditions. Models often wrap their answer in code fences,
commentary or tool transcripts, so the translation is first recovered from the
raw output; Appendix~\ref{sec:extraction} describes this step and how it is
checked.

\begin{table}[t]
\centering
\footnotesize
\setlength{\tabcolsep}{4pt}
\begin{tabular}{@{}l p{3.35cm} r r@{}}
\toprule
Condition & What is added to the request & Tools & Models \\
\midrule
\textsc{zeroshot} & nothing & -- & 31 \\
\textsc{mcp\_prompt} & the tool-use system prompt, with no tools attached (control) & -- & 31 \\
\textsc{inject} & lemma, reading and dictionary entry of every source word, computed in advance, in the user message & 6 & 31 \\
\textsc{inject\_rich} & the same, computed with the thirteen-tool surface & 13 & 14 \\
\textsc{mcp} & the tool-use system prompt, with the tool schemas attached; the model decides whether to call & 6 & 31 \\
\textsc{mcp\_coarse} & as \textsc{mcp}, with one sentence-level tool & 1 & 31 \\
\textsc{mcp\_rich} & as \textsc{mcp}, with the thirteen-tool surface & 13 & 16 \\
\bottomrule
\end{tabular}
\caption{The seven conditions. \emph{Tools} is the surface that supplies the
linguistic material; \emph{Models} is how many of the 31 models ran the
condition for Komi-Zyrian. The other four languages ran \textsc{zeroshot},
\textsc{mcp\_prompt}, \textsc{inject}, \textsc{mcp} and \textsc{mcp\_coarse}
on ten models.}
\label{tab:conditions}
\end{table}

\paragraph{Resources and tool surfaces.} All linguistic information comes
from UralicNLP \citep{hamalainen2019uralicnlp}, which wraps the finite-state
analysers and generators, the constraint grammars and the bilingual
dictionaries of the GiellaLT infrastructure. We expose the library
to the models through three tool surfaces. The \emph{six-tool} surface is the
reference surface of UralicMCP \citep{hamalainen2025uralicmcp}: an analyser,
a lemmatiser, an inflector, a morpheme segmenter, a dictionary lookup and a
list of supported languages, each of which operates on one word. The
\emph{one-tool} surface is a coarse interface we built for this study: it
takes a whole sentence and returns the readings and glosses of every word in
a single call. It tests the hypothesis that composable per-word primitives
suit a programmer but not a small model, which would otherwise have to plan
ten calls where one would do. The \emph{thirteen-tool} surface extends the
six tools with sentence-level glossing, resolution of ISO language codes and
dictionary search by inflected form. The last of these cannot be done on the
per-word surface, which requires the caller to know the lemma already.

\paragraph{Conditions.} Table~\ref{tab:conditions} lists the seven
conditions; the task instruction and the user message are identical in all of
them. Three design decisions need a word of explanation. \textsc{mcp\_prompt}
is a control rather than a system: \textsc{mcp} changes the system prompt and
attaches tools at the same time, so a drop in score under \textsc{mcp} could
be caused by either change, and keeping the prompt while removing the tools
separates the two. \textsc{inject} and \textsc{inject\_rich} contain exactly
the information that the six-tool and the thirteen-tool surfaces return,
computed in advance and written into the user message: every source word with
its lemma, its reading and its dictionary translations. These two conditions
separate the information from the act of fetching it, and they are open to
models that cannot call tools at all.

\paragraph{Models and decoding.} We use open-weight instruction-tuned models
only, with at most 27B parameters. Decoding is greedy with no repetition
penalty, the weights are in bf16, and the context window is 32{,}768 tokens.
The roster is not a full grid. Ten models from 0.6B to 27.8B parameters ran
every direction; a further 21 models ran Komi-Zyrian only and serve the
analysis of tool-calling behaviour in Section~\ref{sec:roster}. Four models
fine-tuned for function calling are included on purpose, since tool calling
is the behaviour under study. All 62{,}920
records are released with their tool-call transcripts and usage
counters,\footnote{The \texttt{runs} subset of the release in \S\ref{sec:data}.} so that Appendix~\ref{sec:extraction} can be checked without a
GPU.

\paragraph{Evaluation.} We evaluate manually, as described in
Section~\ref{sec:error-analysis}. We also computed chrF++ and BLEU for every
record, but we do not use them as evidence: the only reference is a free
literary translation, and on this data the scores mainly reward agreement with
that translator's wording. Appendix~\ref{sec:scores} reports them, together
with the reasons for this decision.

\section{Human evaluation and error analysis}
\label{sec:error-analysis}

An expert linguist read every English output of Qwen3.8-27B against the
reference, sentence by sentence, accepted or rejected it, and assigned one
error category to every rejected output. The analysis covers the 68 source
sentences in each of the four languages under three conditions,
\textsc{zeroshot}, \textsc{inject} and \textsc{mcp}, which gives 816
judgements. We chose this model because it is the only one in the roster
that calls the tools on practically every sentence and still answers in
prose, so it is the only model for which the three conditions really are
three different treatments of the same sentence. What the other models do when
the tools are attached is the subject of Section~\ref{sec:roster}.

An output was accepted when it conveyed the content of the reference sentence:
the same participants, the same event, no invented or omitted material.
Fluency and wording were not criteria, and a loose but faithful rendering was
accepted. Table~\ref{tab:semantic-correctness} reports the result.
\textsc{mcp} has the largest number of accepted outputs (55 of 272), followed
by \textsc{inject} (43) and \textsc{zeroshot} (13). The advantage is not
uniform across languages: \textsc{inject} and \textsc{mcp} tie for Moksha and
Erzya, and \textsc{inject} brings no gain over \textsc{zeroshot} for Udmurt.

The reason Udmurt behaves differently lies in the dictionaries. The
dictionaries in UralicNLP are directional, and the Udmurt dictionary has no
English side. Under
\textsc{inject}, an English gloss reaches the prompt for 80.3\% of the
Komi-Zyrian source tokens, 73.9\% of the Moksha tokens and 78.5\% of the Erzya
tokens, but for none of the Udmurt tokens, whose prompt therefore contains
lemmas and morphological readings only. The gain appears where the glosses
arrive, which separates the effect of the dictionary from the effect of having
any linguistic material in the prompt.

\begin{table}[tbp]
\centering
\small
\renewcommand{\arraystretch}{1.2}
\begin{tabular}{@{}lrrr@{}}
\toprule
Language & \textsc{zeroshot} & \textsc{inject} & \textsc{mcp} \\
\midrule
Komi-Zyrian & 8 (11.8\%) & 17 (25.0\%) & 23 (33.8\%) \\
Moksha & 1 (1.5\%) & 11 (16.2\%) & 11 (16.2\%) \\
Erzya & 0 (0.0\%) & 11 (16.2\%) & 11 (16.2\%) \\
Udmurt & 4 (5.9\%) & 4 (5.9\%) & 10 (14.7\%) \\
\midrule
Total & 13 (4.8\%) & 43 (15.8\%) & 55 (20.2\%) \\
\bottomrule
\end{tabular}
\caption{Outputs accepted as conveying the content of the English reference,
by language and condition. Every output not accepted carries exactly one primary
error category, defined in \S\ref{subsec:taxonomy} and counted in Table~\ref{tab:errorcounts}.}
\label{tab:semantic-correctness}
\end{table}

\subsection{Error taxonomy}
\label{subsec:taxonomy}

Every rejected output received one primary error category from a taxonomy
shared across the four languages. Table~\ref{tab:errorcounts} defines the
categories and gives the counts.

\begin{table*}[t]
\centering
\footnotesize
\setlength{\tabcolsep}{4pt}
\begin{tabular}{@{}l p{6.6cm} r r r r@{}}
\toprule
Primary error & Definition & Komi-Zyrian & Moksha & Erzya & Udmurt \\
\midrule
Wholesale replacement & the situation of the reference is replaced, or several independent major changes are made to it & 39/10/9 & 48/21/26 & 54/24/25 & 42/41/23 \\
Lexical substitution & an action, entity, property or idiomatic meaning is replaced while part of the sentence structure survives & 10/16/15 & 6/12/9 & 9/11/11 & 11/14/13 \\
Participant, referent & who acts or who experiences the event changes: person, gender, kinship or the direction of an interaction & 3/9/9 & 2/8/13 & 1/10/11 & 3/2/7 \\
Polarity, modality, act & negation, propositional force or the status as statement, question or command changes; incompatible event states included & 3/4/3 & 2/5/2 & 0/2/4 & 1/2/4 \\
Omission & an important proposition or semantic component of the reference is missing & 2/7/3 & 8/8/5 & 4/9/5 & 5/4/6 \\
Unsupported addition & a substantive detail that the English reference does not establish is introduced & 2/5/4 & 0/2/2 & 0/1/1 & 1/0/4 \\
Temporal, quantitative & a salient age, duration or temporal relation between events is altered & 0/0/1 & 0/1/0 & 0/0/0 & 0/0/0 \\
Untranslated content & a non-English or name-like form is kept where an English meaning is required & 1/0/1 & 1/0/0 & 0/0/0 & 1/1/1 \\
\midrule
Rejected in total & & 60/51/45 & 67/57/57 & 68/57/57 & 64/64/58 \\
\bottomrule
\end{tabular}
\caption{The error taxonomy, and the primary error category of every rejected
output of Qwen3.8-27B as \textsc{zeroshot}/\textsc{inject}/\textsc{mcp} counts
out of 68 sentences per cell. With the accepted outputs of
Table~\ref{tab:semantic-correctness} every column sums to 68.}
\label{tab:errorcounts}
\end{table*}

Under \textsc{zeroshot}, wholesale replacement is the most frequent category
in every language, with 39, 48, 54 and 42 of the 68 outputs in Komi-Zyrian,
Moksha, Erzya and Udmurt. The injected analyses move most of these outputs
out of that category, to 10, 21 and 24 in the three languages whose dictionary
has an English side, while in Udmurt the count stays at 41. The narrower
categories do not decrease; lexical substitutions and participant errors in
particular become more frequent in every language with an English dictionary
side. In other words, the analyses turn an unrelated output into a nearly
correct one, and the errors that remain are word-level errors, such as a
wrong choice among the glosses of a dictionary entry or a person or possessor
misread from the morphological analysis. The subsections below give one
example for every category that occurs in each language, each headed by the
condition and the sentence identifier.

\subsection{Komi-Zyrian}
\label{subsec:errors-kpv}

\paragraph{Wholesale replacement.} \textsc{zeroshot}, sentence \texttt{neg-006}:
\begin{exquote}
\textbf{Source:}
\begin{otherlanguage}{russian}
Синъясыс пемдісны; горӧдтӧг Татьяна сьӧкыда уси джоджас.
\end{otherlanguage}
\textbf{Output:} The sun had set; without the city, Tatiana felt very lonely.\par
\textbf{Reference:} His vision went dark and without making a sound he fell heavily to the floor.
\end{exquote}
The blackout and collapse in the reference are replaced by sunset and loneliness, changing the setting, participants, and event.

\paragraph{Lexical and conceptual substitution.} \textsc{inject}, sentence \texttt{neg-003}:
\begin{exquote}
\textbf{Source:}
\begin{otherlanguage}{russian}
- Тайӧ дедыд томналіс, - скӧрысь шуис Федя.
\end{otherlanguage}
\textbf{Output:} “Your grandfather was young,” Feda said angrily.\par
\textbf{Reference:} - Grandpa locked it, - Fedya said angrily.
\end{exquote}
The central predicate changes from locking something to being young, even though the speaker and angry delivery are partly preserved.

\paragraph{Participant and referent error.} \textsc{inject}, sentence \texttt{neg-011}:
\begin{exquote}
\textbf{Source:}
\begin{otherlanguage}{russian}
- Ой, ёртъяс, пропадиті!
\end{otherlanguage}
\textbf{Output:} Oh, my friends, you're lost!\par
\textbf{Reference:} - Oh, comrades, I'm done for!
\end{exquote}
The first-person predicament becomes a second-person predicament: the threatened individual is no longer the speaker.

\paragraph{Polarity, modality and speech act.} \textsc{inject}, sentence \texttt{neg-014}:
\begin{exquote}
\textbf{Source:}
\begin{otherlanguage}{russian}
Василий Потупчик лӧгаліс Кулуканов вылӧ да некод водзын эз дзеб ассьыс лӧглунсӧ.
\end{otherlanguage}
\textbf{Output:} Vasily Potupchik was angry at Kulukanov and never showed his anger in front of anyone.\par
\textbf{Reference:} Vasili Potuptshik hated Kulukanov and did not hide his hatred from anyone.
\end{exquote}
The reference explicitly states that the hatred was not concealed, whereas the output says that the anger was never shown. The output also weakens hatred to anger.

\paragraph{Omission.} \textsc{inject}, sentence \texttt{neg-018}:
\begin{exquote}
\textbf{Source:}
\begin{otherlanguage}{russian}
Керкаас вӧлі пӧдтымӧн жар.
\end{otherlanguage}
\textbf{Output:} It was very hot in the house.\par
\textbf{Reference:} The air in the cottage was hot and suffocating.
\end{exquote}
The output retains the hot interior but omits its suffocating quality, an important component of the reference description.

\paragraph{Unsupported addition.} \textsc{mcp}, sentence \texttt{pos-006}:
\begin{exquote}
\textbf{Source:}
\begin{otherlanguage}{russian}
Дона, радейтана мама!
\end{otherlanguage}
\textbf{Output:} Donna, dear mother!\par
\textbf{Reference:} Dear, precious mother!
\end{exquote}
Komi \cyrtext{дона} is the adjective `dear'. Capitalised at the start of the exclamation it looks like a personal name, and the model transliterated it as one, which adds a participant and drops the affectionate description. The same mechanism produces \emph{Musa} below, from \cyrtext{муса} `beloved', and \emph{Kelmoma} in Moksha, from \cyrtext{кельгома} `beloved'.

\paragraph{Temporal and quantitative error.} \textsc{mcp}, sentence \texttt{neg-027}:
\begin{exquote}
\textbf{Source:}
\begin{otherlanguage}{russian}
Эрд вылӧ бӧрдігтырйи котӧрӧн петіс Клава.
\end{otherlanguage}
\textbf{Output:} After crying, Klava came out at a run to the top of the clearing.\par
\textbf{Reference:} Klava ran into the clearing crying.
\end{exquote}
The reference presents crying and running as simultaneous. The output makes the crying precede the running and also adds an unsupported spatial detail.

\paragraph{Untranslated content.} \textsc{zeroshot}, sentence \texttt{pos-007}:
\begin{exquote}
\textbf{Source:}
\begin{otherlanguage}{russian}
Мусаӧй!
\end{otherlanguage}
\textbf{Output:} Musa!\par
\textbf{Reference:} Dear mother!
\end{exquote}
The vocative \cyrtext{Мусаӧй} `my beloved' is read as the name Musa, so the output does not convey a maternal address to an English reader.

\subsection{Moksha}
\label{subsec:errors-mdf}

For Moksha, \textsc{inject} and \textsc{mcp} each yield 11 accepted outputs
against one under \textsc{zeroshot}. The equal totals hide different error
profiles: \textsc{mcp} has more wholesale replacements and more participant
errors than \textsc{inject}, and fewer lexical substitutions and omissions.

\paragraph{Wholesale replacement.} \textsc{zeroshot}, sentence \texttt{neg-025}:
\begin{exquote}
\textbf{Source:}
\begin{otherlanguage}{russian}
Сонь тарсь улалксоц.
\end{otherlanguage}
\textbf{Output:} The sun is setting.\par
\textbf{Reference:} His chin trembled.
\end{exquote}
An event involving a trembling chin is replaced by an unrelated statement about sunset.

\paragraph{Lexical and conceptual substitution.} \textsc{inject}, sentence \texttt{pos-009}:
\begin{exquote}
\textbf{Source:}
\begin{otherlanguage}{russian}
- Цебярьхть каготкятне, Трофим Сергеевич!
\end{otherlanguage}
\textbf{Output:} What lovely coats, Trofim Sergeevich!\par
\textbf{Reference:} - Good papers, Trofim Sergejevitsh.
\end{exquote}
The evaluated objects change from papers to coats; the positive evaluation and the addressee alone do not preserve the proposition.

\paragraph{Participant and referent error.} \textsc{inject}, sentence \texttt{neg-025}:
\begin{exquote}
\textbf{Source:}
\begin{otherlanguage}{russian}
Сонь тарсь улалксоц.
\end{otherlanguage}
\textbf{Output:} She is trembling.\par
\textbf{Reference:} His chin trembled.
\end{exquote}
The output changes the male referent to a female referent and generalizes the trembling from a chin to the whole person.

\paragraph{Polarity, modality and speech act.} \textsc{inject}, sentence \texttt{pos-010}:
\begin{exquote}
\textbf{Source:}
\begin{otherlanguage}{russian}
Аляц рахазевсь:
\end{otherlanguage}
\textbf{Output:} Father was silent:\par
\textbf{Reference:} Father began to laugh.
\end{exquote}
Beginning to laugh is replaced by being silent, producing an incompatible event state.

\paragraph{Omission.} \textsc{inject}, sentence \texttt{neg-022}:
\begin{exquote}
\textbf{Source:}
\begin{otherlanguage}{russian}
Татьянась гневондазь ювади:
\end{otherlanguage}
\textbf{Output:} Tatiana's anger flared up:\par
\textbf{Reference:} Tatjana exclaimed angrily:
\end{exquote}
The emotional state is retained, but the act of exclaiming is omitted.

\paragraph{Unsupported addition.} \textsc{mcp}, sentence \texttt{pos-006}:
\begin{exquote}
\textbf{Source:}
\begin{otherlanguage}{russian}
Кельгома родной тядяй!
\end{otherlanguage}
\textbf{Output:} My dear mother, I love you!\par
\textbf{Reference:} Dear, precious mother!
\end{exquote}
The output adds an explicit declaration of love. Although compatible with the affectionate address, it is not asserted by the reference. This is a flagged borderline rejection under the conservative addition policy.

\paragraph{Temporal and quantitative error.} \textsc{inject}, sentence \texttt{neg-019}:
\begin{exquote}
\textbf{Source:}
\begin{otherlanguage}{russian}
Татьянась ульсь осал и лофтана, эрязень эсь колмогемонь ветие кизонзон фалу трудендазь, радостьта няйсь кржа.
\end{otherlanguage}
\textbf{Output:} Tatiana was thin and pale; in her thirtieth year of life she was always working hard and showed little joy.\par
\textbf{Reference:} Tatjana was thin and pale, she had lived all thirty-five years of her life in constant work. She had seen little joy.
\end{exquote}
The age and duration change from an entire thirty-five-year life of work to work in the thirtieth year. The reference's experience of little joy is also recast as showing little joy.

\paragraph{Untranslated content.} \textsc{zeroshot}, sentence \texttt{pos-006}:
\begin{exquote}
\textbf{Source:}
\begin{otherlanguage}{russian}
Кельгома родной тядяй!
\end{otherlanguage}
\textbf{Output:} My dear Kelmoma!\par
\textbf{Reference:} Dear, precious mother!
\end{exquote}
The adjective \cyrtext{кельгома} `beloved' is read as the name Kelmoma, which replaces the maternal address instead of conveying it.

\subsection{Erzya}
\label{subsec:errors-myv}

No \textsc{zeroshot} output is accepted for Erzya under the
reference-equivalence criterion, against 11 under each of \textsc{inject} and
\textsc{mcp}. This does not mean that every \textsc{zeroshot} output lacked
useful content; partially correct sentences still fail the sentence-level
criterion.

\paragraph{Wholesale replacement.} \textsc{zeroshot}, sentence \texttt{neg-002}:
\begin{exquote}
\textbf{Source:}
\begin{otherlanguage}{russian}
Ледсь ружиясто ды наймунонть прязо лазовсь пельксэнь-пелькс.
\end{otherlanguage}
\textbf{Output:} The cold wind and the rain poured down in sheets.\par
\textbf{Reference:} He fired his shotgun and smashed the merchant's head.
\end{exquote}
The shotgun attack and injury are replaced by a weather description, leaving no equivalent account of the reference event.

\paragraph{Lexical and conceptual substitution.} \textsc{inject}, sentence \texttt{neg-025}:
\begin{exquote}
\textbf{Source:}
\begin{otherlanguage}{russian}
Улозо ноцковтневсь.
\end{otherlanguage}
\textbf{Output:} The chin has been dislocated.\par
\textbf{Reference:} His chin trembled.
\end{exquote}
Trembling is replaced by dislocation, changing the described physical condition.

\paragraph{Participant and referent error.} \textsc{inject}, sentence \texttt{pos-010}:
\begin{exquote}
\textbf{Source:}
\begin{otherlanguage}{russian}
Тетязо пейдезевсь:
\end{otherlanguage}
\textbf{Output:} The aunt laughed.\par
\textbf{Reference:} Father began to laugh.
\end{exquote}
The action of laughing remains, but the father is replaced by an aunt.

\paragraph{Polarity, modality and speech act.} \textsc{mcp}, sentence \texttt{pos-016}:
\begin{exquote}
\textbf{Source:}
\begin{otherlanguage}{russian}
Весе мекс-бути пейдсть.
\end{otherlanguage}
\textbf{Output:} Why are they all laughing?\par
\textbf{Reference:} Everyone was laughing about something.
\end{exquote}
A statement reporting laughter is converted into a question asking for its cause.

\paragraph{Omission.} \textsc{inject}, sentence \texttt{neg-022}:
\begin{exquote}
\textbf{Source:}
\begin{otherlanguage}{russian}
Татьяна кежейстэ рангстась:
\end{otherlanguage}
\textbf{Output:} Tatiana was furious.\par
\textbf{Reference:} Tatjana exclaimed angrily:
\end{exquote}
The output expresses anger but omits the speech event specified by exclaimed.

\paragraph{Unsupported addition.} \textsc{mcp}, sentence \texttt{neg-005}:
\begin{exquote}
\textbf{Source:}
\begin{otherlanguage}{russian}
Куростонть маряви авань кежей рангома:
\end{otherlanguage}
\textbf{Output:} The mother's angry scream was heard from the grove:\par
\textbf{Reference:} From outside came an angry woman's voice:
\end{exquote}
The output narrows the unspecified woman to a mother and outside to a grove. These details are unsupported by the English reference; source-language adjudication would be needed to determine whether they are genuine translation errors.

\subsection{Udmurt}
\label{subsec:errors-udm}

For Udmurt, \textsc{zeroshot} and \textsc{inject} each yield four accepted
outputs and \textsc{mcp} ten. \textsc{inject} therefore brings no gain in the
one language whose prompt carries no English glosses. \textsc{mcp} halves the
number of wholesale replacements, while lexical, participant and omission
errors remain.

\paragraph{Wholesale replacement.} \textsc{zeroshot}, sentence \texttt{pos-007}:
\begin{exquote}
\textbf{Source:}
\begin{otherlanguage}{russian}
Дуное!
\end{otherlanguage}
\textbf{Output:} Damn it!\par
\textbf{Reference:} Dear mother!
\end{exquote}
An affectionate maternal address becomes profanity, replacing both the referent and the communicative function.

\paragraph{Lexical and conceptual substitution.} \textsc{mcp}, sentence \texttt{pos-010}:
\begin{exquote}
\textbf{Source:}
\begin{otherlanguage}{russian}
Атай серекъяны ӧдъяз:
\end{otherlanguage}
\textbf{Output:} Father began to mock:\par
\textbf{Reference:} Father began to laugh.
\end{exquote}
Laughing becomes mocking, adding a derisive meaning that the reference does not establish.

\paragraph{Participant and referent error.} \textsc{mcp}, sentence \texttt{neg-020}:
\begin{exquote}
\textbf{Source:}
\begin{otherlanguage}{russian}
Пинал дыръяз солы дышетскыны ӧз кылды: узыр бускельзы дорын медъяськыса ужалляз.
\end{otherlanguage}
\textbf{Output:} In his youth, he was unable to study: he was hired to work at the home of a wealthy neighbor.\par
\textbf{Reference:} As a child she had had no opportunity to study: she had worked as a servant for a wealthy neighbor.
\end{exquote}
The reference's female participant becomes male. Much of the education and employment content survives, but the participant identity is not equivalent.

\paragraph{Polarity, modality and speech act.} \textsc{inject}, sentence \texttt{neg-033}:
\begin{exquote}
\textbf{Source:}
\begin{otherlanguage}{russian}
Одӥгезлэн кышкамез мукетъёссэ куалектытэ:
\end{otherlanguage}
\textbf{Output:} One person's fear does not affect the others:\par
\textbf{Reference:} Cowardice is contagious.
\end{exquote}
The proposition that cowardice spreads is reversed into a claim that one person's fear does not affect others.

\paragraph{Omission.} \textsc{inject}, sentence \texttt{neg-005}:
\begin{exquote}
\textbf{Source:}
\begin{otherlanguage}{russian}
Урамысь лек нылкышно куара кылӥськиз:
\end{otherlanguage}
\textbf{Output:} A woman's voice was heard from the street:\par
\textbf{Reference:} From outside came an angry woman's voice:
\end{exquote}
The location and the woman's voice remain, but the angry quality is omitted.

\paragraph{Unsupported addition.} \textsc{mcp}, sentence \texttt{neg-008}:
\begin{exquote}
\textbf{Source:}
\begin{otherlanguage}{russian}
Песятай вожез потыса киосыныз шона, шиетэ:
\end{otherlanguage}
\textbf{Output:} The old man, having come out of the river, shook his hands and hissed:\par
\textbf{Reference:} Grandpa waved his hands furiously and hissed:
\end{exquote}
The output introduces a prior event in which the old man emerges from a river, which is absent from the reference. It also omits the explicitly furious manner.

\paragraph{Untranslated content.} \textsc{mcp}, sentence \texttt{pos-028}:
\begin{exquote}
\textbf{Source:}
\begin{otherlanguage}{russian}
- Э! Кампет! - шумпотыса черекъя Федя.
\end{otherlanguage}
\textbf{Output:} - Hey! Kampet! - Fedya called out, rejoicing.\par
\textbf{Reference:} - Candies! - Fedja squealed happily.
\end{exquote}
The common noun \cyrtext{кампет} `candies', capitalised after the interjection, is kept as the name-like form Kampet. The joyful delivery does not compensate for the untranslated content word.

\section{Tool calling across the roster}
\label{sec:roster}

The manual evaluation covers one model. What did the other 30 models do when
the tools were attached? That question concerns the behaviour of the models
rather than the quality of their translations, and it can be answered by
counting.

\begin{table}[tb]
\centering\scriptsize
\setlength{\tabcolsep}{3pt}
\begin{tabular}{@{}l r r@{\quad}l r r@{}}
\toprule
Model & Size & Calls & Model & Size & Calls \\
\midrule
Qwen3-0.6B & 0.6 & 0\% & Mistral-7B-Instruct-v0.3 & 7.0 & 0\% \\
gemma-3-1b & 1.0 & 0\% & OLMo-2-1124-7B & 7.0 & 0\% \\
xLAM-2-1b & 1.0 & 23\% & Qwen2.5-7B & 7.0 & 3\% \\
Llama-3.2-1B & 1.2 & 71\% & Hermes-3-Llama-3.1-8B & 8.0 & 7\% \\
Hammer2.1-1.5b & 1.5 & 0\% & Llama-3.1-8B & 8.0 & 98\% \\
Qwen2.5-1.5B & 1.5 & 50\% & Ministral-8B & 8.0 & 100\% \\
EuroLLM-1.7B & 1.7 & 0\% & Qwen3-8B & 8.0 & 46\% \\
Qwen3-1.7B & 1.7 & 97\% & c4ai-command-r7b-12-2024 & 8.0 & 0\% \\
granite-3.3-2b & 2.0 & 42\% & granite-3.3-8b & 8.0 & $<$1\% \\
Hammer2.1-3b & 3.0 & 0\% & watt-tool-8B & 8.0 & 0\% \\
Qwen2.5-3B & 3.0 & 58\% & EuroLLM-9B & 9.0 & 0\% \\
xLAM-2-3b & 3.0 & 100\% & Mistral-Nemo & 12.0 & 100\% \\
Llama-3.2-3B & 3.2 & 85\% & gemma-3-12b & 12.0 & 0\% \\
Phi-4-mini & 3.8 & 0\% & gemma-3-27b & 27.0 & 0\% \\
Qwen3-4B & 4.0 & 96\% & Qwen3.8-27B & 27.8 & 100\% \\
gemma-3-4b & 4.0 & 0\% &  & &  \\
\bottomrule
\end{tabular}
\caption{Share of sentences on which each model emitted at least one tool call under \textsc{mcp}, \textsc{mcp\_coarse} and \textsc{mcp\_rich}. Fourteen models never call, and one more calls on fewer than one sentence in two hundred. \emph{Size} is in billions of parameters.}
\label{tab:callrate}
\end{table}

\paragraph{Call rates.} Attaching a tool does not make a model use it. Whether
to call is the model's own decision, taken sentence by sentence.
Table~\ref{tab:callrate} gives the share of sentences on which each model
emitted at least one tool call under the three \textsc{mcp} conditions. Across
the study, 57.6\% of the records in a condition with tools attached contain no
call, and 14 of the 31 models never call at all. For these models the
\textsc{mcp} condition is the \textsc{zeroshot} condition with a longer system
prompt. Model size does not predict the call rate. Gemma never calls
at 1B, 4B, 12B or 27B parameters, Qwen3 calls at every size from 1.7B upwards,
and Mistral-7B never calls while Ministral-8B always does. The behaviour is
stable within a model family and ranges from 0\% to 100\% across families of
the same size.

\paragraph{Capability and willingness.} A model that never calls may be unable
to produce a usable call or may choose not to. Appendix~\ref{sec:probe}
separates the two cases with three tests: a direct instruction to call, an
exhaustive search over the tool-call parsers of the serving framework, and the
sweep itself. Nine of the 31 models fail all three tests, and two more emit
call syntax that no parser reads; for these eleven models the call format is
absent rather than unused. The remaining three silent models are able to call
and do not. All our conditions are zero-shot, and a worked example of a tool
call might change the behaviour of these three models, which we have not
tested.

\paragraph{What the calls contain.} Every call names the language to be
processed by an ISO code, and most of these codes are wrong. Of the 315{,}910
codes sent during the study, 42.5\% name the language actually being
processed, 49.1\% name no language that UralicNLP supports, and 8.4\% name a
supported but wrong language; 26{,}630 calls ask for \texttt{koi},
Komi-Permyak, on Komi-Zyrian text. As a consequence, 52.4\% of all calls return
an empty result. Appendix~\ref{sec:isocodes} gives these figures per model.
The injected prompt, by contrast, never has to name the language and delivers
the same material for every sentence.

\paragraph{Function-calling specialists.} Four models in the roster are
fine-tuned for function calling. They call readily, but they answer in call
syntax rather than in prose, so that often no translation can be found in
their output. This is the case for 42.4\% of the outputs of Hammer2.1-3b and
17.2\% of the outputs of xLAM-2-1b-fc-r, against 4.3\% for the models that are
not specialised in this way.

\paragraph{Translation into the endangered languages.}
Table~\ref{tab:pathology} in Appendix~\ref{sec:failuremodes} classifies every
output in the study by its form. Out of the four languages, 98.1\% of the
zero-shot outputs are well-formed English. Into them, only 72.0\% of the
zero-shot outputs are well-formed text in the target language, and the share
falls to 42.5\% under \textsc{mcp}, with 22.9\% of the outputs in the wrong
language and 13.1\% caught in a repetition loop under greedy decoding. The
injected analyses keep the share at 67.1\% and change nothing else. The
rule-based resources supply lemmas, readings and glosses; what is missing in
this direction is a model of the target language itself, which neither an
analyser nor a dictionary can provide.

\section{Conclusions}
\label{sec:discussion}

Our main result is that rule-based analyses written into the prompt improve
translation from Komi-Zyrian, Moksha and Erzya into English. On the model we evaluated
manually, the number of accepted translations rises from 13 to 43 out of 272,
and the remaining errors shift from wholesale replacement of the sentence to
word-level substitutions. The gain depends on the bilingual dictionary and is
absent for Udmurt, whose dictionary has no English side. The same analyses
delivered as tools over MCP help the one model that calls the tools reliably
and answers in prose, with 55 accepted translations, but we cannot recommend
this route in general: 14 of the 31 models never call, 11 of them cannot, and
half of the calls that are made name a language the toolkit does not know.
Translation into the four languages fails under every condition, because the
resources do not supply a model of the target language.

These conclusions rest on a narrow basis. The manual judgements cover one
model, one annotator and 68 sentences from one text, and the comparison
between the two delivery routes rests on that model alone. Whether the
advantage of the tools over the injected analyses holds for other models that
call reliably is a question we cannot answer with this data.

\section*{Limitations}

The test set consists of 68 sentences from one literary text. The English
reference is a free translation made independently of the Uralic ones, which
is why we evaluate manually rather than against the reference. The manual
judgements come from one annotator and cover one model, so they describe the
best case in the roster rather than a typical member of it. Recovering the
translation from each raw output relies on a language model; the step is
checked by verbatim containment on every record, but not by a person.

The test set is balanced for sentiment polarity, with a gold label on every
sentence, because a translation can match its reference closely and still
invert the emotional tone of the source. Our evaluation does not use these
labels. Measuring polarity preservation would require a sentiment classifier
for the target language, and none exists for Komi-Zyrian, Moksha, Erzya or
Udmurt; the alternative is annotation by an expert linguist, which is what
Section~\ref{sec:error-analysis} does for one model and which does not scale
to seven conditions. Until such a measure exists, our evaluation cannot
distinguish a system that conveys the content of a sentence from one that
also conveys its tone.

The four languages tokenise badly in every model of the roster, at 3.50 to
4.98 subword tokens per Komi-Zyrian word against 1.36 to 1.59 per English word
(Table~\ref{tab:fertility} in Appendix~\ref{sec:fertility}). This does not
affect the comparison between conditions, since the analyses reach the model
through the same tokeniser in every condition, but it means that the same
context window holds two to three times less text in an endangered language,
which neither a better prompt nor a better tool can change.

All models ran in bf16, whereas a local deployment typically runs 4-bit or
8-bit weights. Emitting a well-formed tool call is a format-sensitive
behaviour that quantisation is known to disturb early
\citep{siniaev2026compressed}, so the call rates reported here are an
optimistic estimate of what a deployed system would show.

\bibliography{custom}

@article{hamalainen2019uralicnlp,
  title = {{UralicNLP}: An {NLP} Library for {U}ralic Languages},
  author = {H{\"a}m{\"a}l{\"a}inen, Mika},
  journal = {Journal of Open Source Software},
  volume = {4}, number = {37}, pages = {1345}, year = {2019},
  doi = {10.21105/joss.01345}
}

@article{rueter2021prerequisites,
	doi = {10.20944/preprints202104.0131.v1},
	url = {https://doi.org/10.20944/preprints202104.0131.v1},
	year = 2021,
	month = {April},
	publisher = {Preprints},
	author = {Jack Rueter and Mika Hämäläinen},
	title = {Prerequisites for Shallow-Transfer Machine Translation of Mordvin Languages: Language Documentation with a Purpose},
	journal = {Preprints}
}

@inproceedings{yankovskaya2023machine,
    title = "Machine Translation for Low-resource {F}inno-{U}gric Languages",
    author = {Yankovskaya, Lisa  and
      Tars, Maali  and
      T{\"a}ttar, Andre  and
      Fishel, Mark},
    editor = {Alum{\"a}e, Tanel  and
      Fishel, Mark},
    booktitle = "Proceedings of the 24th {N}ordic Conference on Computational Linguistics ({N}o{D}a{L}i{D}a)",
    month = may,
    year = "2023",
    address = "T{\'o}rshavn, Faroe Islands",
    publisher = "University of Tartu Library",
    url = "https://aclanthology.org/2023.nodalida-1.77/",
    pages = "762--771"
}

@inproceedings{dale2022first,
    title = "The first neural machine translation system for the {E}rzya language",
    author = "Dale, David",
    editor = "Serikov, Oleg  and
      Voloshina, Ekaterina  and
      Postnikova, Anna  and
      Klyachko, Elena  and
      Neminova, Ekaterina  and
      Vylomova, Ekaterina  and
      Shavrina, Tatiana  and
      Ferrand, Eric Le  and
      Malykh, Valentin  and
      Tyers, Francis  and
      Arkhangelskiy, Timofey  and
      Mikhailov, Vladislav  and
      Fenogenova, Alena",
    booktitle = "Proceedings of the First Workshop on {NLP} applications to field linguistics",
    month = oct,
    year = "2022",
    address = "Gyeongju, Republic of Korea",
    publisher = "International Conference on Computational Linguistics",
    url = "https://aclanthology.org/2022.fieldmatters-1.6/",
    pages = "45--53"
}

@inproceedings{alnajjar2023bootstrapping,
    title = "Bootstrapping {M}oksha-{E}rzya Neural Machine Translation from Rule-Based Apertium",
    author = {Alnajjar, Khalid  and
      H{\"a}m{\"a}l{\"a}inen, Mika  and
      Rueter, Jack},
    editor = {H{\"a}m{\"a}l{\"a}inen, Mika  and
      {\"O}hman, Emily  and
      Pirinen, Flammie  and
      Alnajjar, Khalid  and
      Miyagawa, So  and
      Bizzoni, Yuri  and
      Partanen, Niko  and
      Rueter, Jack},
    booktitle = "Proceedings of the Joint 3rd International Conference on Natural Language Processing for Digital Humanities and 8th International Workshop on Computational Linguistics for {U}ralic Languages",
    month = dec,
    year = "2023",
    address = "Tokyo, Japan",
    publisher = "Association for Computational Linguistics",
    url = "https://aclanthology.org/2023.nlp4dh-1.24/",
    pages = "213--218"
}

@inproceedings{johnson2017north,
    title = "{N}orth-{S}{\'a}mi to {F}innish rule-based machine translation system",
    author = "Johnson, Ryan  and
      Pirinen, Tommi A  and
      Puolakainen, Tiina  and
      Tyers, Francis  and
      Trosterud, Trond  and
      Unhammer, Kevin",
    editor = {Tiedemann, J{\"o}rg  and
      Tahmasebi, Nina},
    booktitle = "Proceedings of the 21st {N}ordic Conference on Computational Linguistics",
    month = may,
    year = "2017",
    address = "Gothenburg, Sweden",
    publisher = "Association for Computational Linguistics",
    url = "https://aclanthology.org/W17-0214/",
    pages = "115--122"
}

@inproceedings{pirinen2022building,
    title = "Building an Extremely Low Resource Language to High Resource Language Machine Translation System from Scratch",
    author = "Pirinen, Flammie A  and
      Wiechetek, Linda",
    editor = "Schaefer, Robin  and
      Bai, Xiaoyu  and
      Stede, Manfred  and
      Zesch, Torsten",
    booktitle = "Proceedings of the 18th Conference on Natural Language Processing ({KONVENS} 2022)",
    month = "12--15 " # sep,
    year = "2022",
    address = "Potsdam, Germany",
    publisher = "KONVENS 2022 Organizers",
    url = "https://aclanthology.org/2022.konvens-1.18/",
    pages = "150--155"
}

@inproceedings{alnajjar-etal-2023-sentiment,
    title = "Sentiment Analysis Using Aligned Word Embeddings for {U}ralic Languages",
    author = {Alnajjar, Khalid  and
      H{\"a}m{\"a}l{\"a}inen, Mika  and
      Rueter, Jack},
    editor = "Ilinykh, Nikolai  and
      Morger, Felix  and
      Dann{\'e}lls, Dana  and
      Dobnik, Simon  and
      Megyesi, Be{\'a}ta  and
      Nivre, Joakim",
    booktitle = "Proceedings of the Second Workshop on Resources and Representations for Under-Resourced Languages and Domains ({RESOURCEFUL}-2023)",
    month = may,
    year = "2023",
    address = "T{\'o}rshavn, the Faroe Islands",
    publisher = "Association for Computational Linguistics",
    url = "https://aclanthology.org/2023.resourceful-1.3/",
    pages = "19--24"
}

@inproceedings{popovic2017chrfpp,
    title = "chr{F}++: words helping character n-grams",
    author = "Popovi{\'c}, Maja",
    editor = "Bojar, Ond{\v{r}}ej  and
      Buck, Christian  and
      Chatterjee, Rajen  and
      Federmann, Christian  and
      Graham, Yvette  and
      Haddow, Barry  and
      Huck, Matthias  and
      Yepes, Antonio Jimeno  and
      Koehn, Philipp  and
      Kreutzer, Julia",
    booktitle = "Proceedings of the Second Conference on Machine Translation",
    month = sep,
    year = "2017",
    address = "Copenhagen, Denmark",
    publisher = "Association for Computational Linguistics",
    url = "https://aclanthology.org/W17-4770/",
    doi = "10.18653/v1/W17-4770",
    pages = "612--618"
}

@inproceedings{papineni2002bleu,
    title = "{B}leu: a Method for Automatic Evaluation of Machine Translation",
    author = "Papineni, Kishore  and
      Roukos, Salim  and
      Ward, Todd  and
      Zhu, Wei-Jing",
    editor = "Isabelle, Pierre  and
      Charniak, Eugene  and
      Lin, Dekang",
    booktitle = "Proceedings of the 40th Annual Meeting of the {A}ssociation for {C}omputational {L}inguistics",
    month = jul,
    year = "2002",
    address = "Philadelphia, Pennsylvania, USA",
    publisher = "Association for Computational Linguistics",
    url = "https://aclanthology.org/P02-1040/",
    doi = "10.3115/1073083.1073135",
    pages = "311--318"
}

@inproceedings{post2018sacrebleu,
    title = "A Call for Clarity in Reporting {BLEU} Scores",
    author = "Post, Matt",
    editor = "Bojar, Ond{\v{r}}ej  and
      Chatterjee, Rajen  and
      Federmann, Christian  and
      Fishel, Mark  and
      Graham, Yvette  and
      Haddow, Barry  and
      Huck, Matthias  and
      Yepes, Antonio Jimeno  and
      Koehn, Philipp  and
      Monz, Christof  and
      Negri, Matteo  and
      N{\'e}v{\'e}ol, Aur{\'e}lie  and
      Neves, Mariana  and
      Post, Matt  and
      Specia, Lucia  and
      Turchi, Marco  and
      Verspoor, Karin",
    booktitle = "Proceedings of the Third Conference on Machine Translation: Research Papers",
    month = oct,
    year = "2018",
    address = "Brussels, Belgium",
    publisher = "Association for Computational Linguistics",
    url = "https://aclanthology.org/W18-6319/",
    doi = "10.18653/v1/W18-6319",
    pages = "186--191"
}

@inproceedings{koehn2004significance,
    title = "Statistical Significance Tests for Machine Translation Evaluation",
    author = "Koehn, Philipp",
    editor = "Lin, Dekang  and
      Wu, Dekai",
    booktitle = "Proceedings of the 2004 Conference on Empirical Methods in Natural Language Processing",
    month = jul,
    year = "2004",
    address = "Barcelona, Spain",
    publisher = "Association for Computational Linguistics",
    url = "https://aclanthology.org/W04-3250/",
    pages = "388--395"
}

@misc{mcp2024,
  title = {Model Context Protocol},
  author = {{Anthropic}}, year = {2024},
  howpublished = {\url{https://modelcontextprotocol.io}}
}

@inproceedings{rust2021tokenizer,
    title = "How Good is Your Tokenizer? On the Monolingual Performance of Multilingual Language Models",
    author = "Rust, Phillip  and
      Pfeiffer, Jonas  and
      Vuli{\'c}, Ivan  and
      Ruder, Sebastian  and
      Gurevych, Iryna",
    editor = "Zong, Chengqing  and
      Xia, Fei  and
      Li, Wenjie  and
      Navigli, Roberto",
    booktitle = "Proceedings of the 59th Annual Meeting of the {A}ssociation for {C}omputational {L}inguistics and the 11th International Joint Conference on Natural Language Processing (Volume 1: Long Papers)",
    month = aug,
    year = "2021",
    address = "Online",
    publisher = "Association for Computational Linguistics",
    url = "https://aclanthology.org/2021.acl-long.243/",
    doi = "10.18653/v1/2021.acl-long.243",
    pages = "3118--3135"
}

@inproceedings{hamalainen2025uralicmcp,
  title = {{UralicMCP}: Turning {LLM}s into Experts in Endangered Languages with {MCP}},
  author = {H{\"a}m{\"a}l{\"a}inen, Mika and Rueter, Jack},
  booktitle = {Lightning Proceedings of the 10th International Workshop on
               Computational Linguistics for Uralic Languages},
  year = {2025},
  url = {https://hal.science/hal-05406638v1/file/UralicMCP.pdf}
}

@inproceedings{rueter2020mordvinic,
  title = {Open-Source Morphology for Endangered {M}ordvinic Languages},
  author = {Rueter, Jack and H{\"a}m{\"a}l{\"a}inen, Mika and Partanen, Niko},
  booktitle = {Proceedings of Second Workshop for {NLP} Open Source Software
               ({NLP-OSS})},
  pages = {94--100}, year = {2020},
  doi = {10.18653/v1/2020.nlposs-1.13}
}

@inproceedings{hamalainen2024dag,
  title = {{DAG}: Dictionary-Augmented Generation for Disambiguation of
           Sentences in Endangered {U}ralic Languages using {C}hat{GPT}},
  author = {H{\"a}m{\"a}l{\"a}inen, Mika},
  booktitle = {Proceedings of the 9th International Workshop on Computational
               Linguistics for Uralic Languages},
  year = {2024},
  url = {https://arxiv.org/abs/2411.01531}
}

@inbook{hamalainen2021endangered,
   title={Endangered Languages are not Low-Resourced!},
   ISBN={9789515150257},
   url={http://dx.doi.org/10.31885/9789515150257.1},
   DOI={10.31885/9789515150257.1},
   booktitle={Multilingual Facilitation},
   publisher={University of Helsinki},
   author={Hämäläinen, Mika},
   year={2021},
   month=Mar, pages={1–11} 
}

@inproceedings{moshagen2023giellalt,
  title = {The {G}iella{LT} Infrastructure: A Multilingual Infrastructure for
           Rule-Based {NLP}},
  author = {Moshagen, Sjur N{\o}rsteb{\o} and Pirinen, Flammie and Antonsen,
            Lene and Gaup, B{\o}rre and Mikkelsen, Inga and Trosterud, Trond
            and Wiechetek, Linda and Hiovain-Asikainen, Katri},
  booktitle = {Proceedings of the 24th Nordic Conference on Computational
               Linguistics (NoDaLiDa)}, year = {2023}
}

@inproceedings{rueter2021komi,
    title = "Overview of Open-Source Morphology Development for the {K}omi-{Z}yrian Language: Past and future",
    author = {Rueter, Jack  and
      Partanen, Niko  and
      H{\"a}m{\"a}l{\"a}inen, Mika  and
      Trosterud, Trond},
    editor = "Pirinen, Flammie A  and
      Arhangelskiy, Timofey  and
      Trosterud, Trond  and
      Rie{\ss}ler, Michael",
    booktitle = "Proceedings of the Seventh International Workshop on Computational Linguistics of {U}ralic Languages",
    month = sep,
    year = "2021",
    address = "Syktyvkar, Russia (Online)",
    publisher = "Association for Computational Linguistics",
    url = "https://aclanthology.org/2021.iwclul-1.4/",
    pages = "29--39"
}

@inproceedings{tereschenko2025gpt,
    title = "Evaluating {O}pen{AI} {GPT} Models for Translation of Endangered {U}ralic{L}anguages: A Comparison of Reasoning and Non-Reasoning Architectures",
    author = {Tereshchenko, Yehor  and
      H{\"a}m{\"a}l{\"a}inen, Mika  and
      Myroniuk, Svitlana},
    editor = {H{\"a}m{\"a}l{\"a}inen, Mika  and
      Rie{\ss}ler, Michael  and
      Morooka, Eiaki V.  and
      Kharlashkin, Lev},
    booktitle = "Proceedings of the 10th International Workshop on Computational Linguistics for {U}ralic Languages",
    month = dec,
    year = "2025",
    address = "Joensuu, Finland",
    publisher = "Association for Computational Linguistics",
    url = "https://aclanthology.org/2025.iwclul-1.17/",
    pages = "131--139",
    ISBN = "979-8-89176-360-9"
}

@inproceedings{xu2026tokenization,
    title = "Tokenization and Morphological Fidelity in {U}ralic {NLP}: A Cross-Lingual Evaluation",
    author = "Xu, Nuo  and
      Kim, Ahrii",
    editor = "Hettiarachchi, Hansi  and
      Ranasinghe, Tharindu  and
      Plum, Alistair  and
      Rayson, Paul  and
      Mitkov, Ruslan  and
      Gaber, Mohamed  and
      Premasiri, Damith  and
      Tan, Fiona Anting  and
      Uyangodage, Lasitha",
    booktitle = "Proceedings of the Second Workshop on Language Models for Low-Resource Languages ({L}o{R}es{LM} 2026)",
    month = mar,
    year = "2026",
    address = "Rabat, Morocco",
    publisher = "Association for Computational Linguistics",
    url = "https://aclanthology.org/2026.loreslm-1.43/",
    doi = "10.18653/v1/2026.loreslm-1.43",
    pages = "493--503",
    ISBN = "979-8-89176-377-7"
}

@inproceedings{chelombitko2024tokenizers,
    title = "Specialized Monolingual {BPE} Tokenizers for {U}ralic Languages Representation in Large Language Models",
    author = "Chelombitko, Iaroslav  and
      Komissarov, Aleksey",
    editor = {H{\"a}m{\"a}l{\"a}inen, Mika  and
      Pirinen, Flammie  and
      Macias, Melany  and
      Crespo Avila, Mario},
    booktitle = "Proceedings of the 9th International Workshop on Computational Linguistics for {U}ralic Languages",
    month = nov,
    year = "2024",
    address = "Helsinki, Finland",
    publisher = "Association for Computational Linguistics",
    url = "https://aclanthology.org/2024.iwclul-1.11/",
    pages = "89--95"
}

@inproceedings{purason2025finnougric,
    title = "{LLM}s for Extremely Low-Resource {F}inno-{U}gric Languages",
    author = "Purason, Taido  and
      Kuulmets, Hele-Andra  and
      Fishel, Mark",
    editor = "Chiruzzo, Luis  and
      Ritter, Alan  and
      Wang, Lu",
    booktitle = "Findings of the {A}ssociation for {C}omputational {L}inguistics: {NAACL} 2025",
    month = apr,
    year = "2025",
    address = "Albuquerque, New Mexico",
    publisher = "Association for Computational Linguistics",
    url = "https://aclanthology.org/2025.findings-naacl.373/",
    doi = "10.18653/v1/2025.findings-naacl.373",
    pages = "6692--6712",
    ISBN = "979-8-89176-195-7"
}

@inproceedings{chelombitko2025samponlp,
    title = "{S}ampo{NLP}: A Self-Referential Toolkit for Morphological Analysis of Subword Tokenizers",
    author = "Chelombitko, Iaroslav  and
      Chelombitko, Ekaterina  and
      Komissarov, Aleksey",
    editor = {H{\"a}m{\"a}l{\"a}inen, Mika  and
      Rie{\ss}ler, Michael  and
      Morooka, Eiaki V.  and
      Kharlashkin, Lev},
    booktitle = "Proceedings of the 10th International Workshop on Computational Linguistics for {U}ralic Languages",
    month = dec,
    year = "2025",
    address = "Joensuu, Finland",
    publisher = "Association for Computational Linguistics",
    url = "https://aclanthology.org/2025.iwclul-1.8/",
    pages = "57--67",
    ISBN = "979-8-89176-360-9"
}

@inproceedings{rei2020comet,
    title = "{COMET}: A Neural Framework for {MT} Evaluation",
    author = "Rei, Ricardo  and
      Stewart, Craig  and
      Farinha, Ana C  and
      Lavie, Alon",
    editor = "Webber, Bonnie  and
      Cohn, Trevor  and
      He, Yulan  and
      Liu, Yang",
    booktitle = "Proceedings of the 2020 Conference on Empirical Methods in Natural Language Processing ({EMNLP})",
    month = nov,
    year = "2020",
    address = "Online",
    publisher = "Association for Computational Linguistics",
    url = "https://aclanthology.org/2020.emnlp-main.213/",
    doi = "10.18653/v1/2020.emnlp-main.213",
    pages = "2685--2702"
}

@misc{siniaev2026compressed,
      title={Compressed code: the hidden effects of quantization and distillation on programming tokens}, 
      author={Viacheslav Siniaev and Iaroslav Chelombitko and Aleksey Komissarov},
      year={2026},
      eprint={2601.02563},
      archivePrefix={arXiv},
      primaryClass={cs.SE},
      url={https://arxiv.org/abs/2601.02563}, 
}

\clearpage
\appendix
\section*{Appendix}
\nopagebreak

\section{Automatic scores, for reference}
\label{sec:scores}

We computed chrF++ \citep{popovic2017chrfpp} and BLEU \citep{papineni2002bleu},
with sacreBLEU \citep{post2018sacrebleu},\footnote{sacreBLEU signature
\texttt{nrefs:1\allowbreak |case:mixed\allowbreak |eff:no\allowbreak
|tok:13a\allowbreak |smooth:exp}.} for every record, and chrF++ on text
lemmatised with the same FSTs on both sides for the directions into the Uralic
languages. We do not use these scores as evidence in the paper, for two
reasons. First, the only reference is a free literary translation, so a score
mainly rewards agreement with that translator's wording. Second, trained
metrics such as COMET \citep{rei2020comet} require a pretrained encoder for the
target language, and none exists for these four languages. We report the
scores here so that readers can compare with other work. Every score should be
read against a \emph{floor}, the same metric computed on the references paired
with the wrong sentence: 11.5 chrF++ for English targets and 9.3 to 10.8 for
the Uralic ones. Character overlap rewards any string in the right alphabet,
and a score below the floor means that the output resembles the reference
less than an unrelated sentence does.

\begin{table*}[t]
\centering
\footnotesize
\setlength{\tabcolsep}{4pt}
\begin{tabular}{l r r r r r}
\toprule
Direction & no tools & tool prompt & inject & mcp & mcp coarse \\
\midrule
Komi-Zyrian$\to$English & 19.8 & 15.6 (-4.3) & 25.3 (+5.5) & 18.5 (-1.3) & 18.9 (-1.0) \\
Moksha$\to$English & 17.4 & 16.0 (-1.4) & 22.1 (+4.7) & 15.2 (-2.2) & 15.2 (-2.2) \\
Erzya$\to$English & 15.1 & 11.9 (-3.2) & 20.0 (+4.9) & 14.5 (-0.6) & 14.5 (-0.6) \\
Udmurt$\to$English & 16.3 & 14.0 (-2.2) & 16.4 (+0.2) & 14.8 (-1.4) & 15.6 (-0.7) \\
English$\to$Komi-Zyrian & 4.4 & 3.3 (-1.1) & 3.7 (-0.7) & 4.5 (+0.1) & 2.6 (-1.8) \\
\midrule
English$\to$Moksha & 3.4 & 2.3 (-1.1) & 2.1 (-1.4) & 3.5 (+0.1) & 2.0 (-1.5) \\
English$\to$Erzya & 4.0 & 3.3 (-0.7) & 2.9 (-1.0) & 4.4 (+0.5) & 2.7 (-1.2) \\
English$\to$Udmurt & 5.4 & 2.9 (-2.4) & 4.0 (-1.3) & 3.1 (-2.3) & 3.7 (-1.7) \\
\bottomrule
\end{tabular}
\caption{chrF++ averaged over the ten models that ran every direction, 68 sentences each. Parentheses give the change from the zero-shot baseline. The floor is 11.5 for English targets and 9.3 to 10.8 for the Uralic ones.}
\label{tab:crosslang}
\end{table*}

\begin{table}[tb]
\centering\footnotesize
\setlength{\tabcolsep}{4pt}
\begin{tabular}{@{}l rr rr rr@{}}
\toprule
& \multicolumn{2}{c}{\textsc{zeroshot}} & \multicolumn{2}{c}{\textsc{inject}} & \multicolumn{2}{c}{\textsc{mcp}} \\
\cmidrule(lr){2-3}\cmidrule(lr){4-5}\cmidrule(lr){6-7}
Language & acc. & chrF & acc. & chrF & acc. & chrF \\
\midrule
Komi-Zyrian & 8 & 29.9 & 17 & 37.8 & 23 & 40.5 \\
Moksha & 1 & 22.4 & 11 & 30.2 & 11 & 31.6 \\
Erzya & 0 & 20.2 & 11 & 29.7 & 11 & 30.9 \\
Udmurt & 4 & 24.5 & 4 & 26.2 & 10 & 29.7 \\
\bottomrule
\end{tabular}
\caption{The twelve cells where both kinds of evidence exist: Qwen3.8-27B, four languages, three conditions. \emph{acc.} is how many of the 68 outputs an expert linguist accepted; \emph{chrF} is chrF++ on the same outputs. The two agree at $r=0.98$ (Pearson) and $\rho=0.90$ (Spearman).}
\label{tab:humanauto}
\end{table}

Table~\ref{tab:crosslang} gives chrF++ per direction and condition, averaged
over the ten models that ran every direction. Table~\ref{tab:humanauto} puts
the scores beside the manual judgements on the twelve cells where both exist.
On these cells the two agree at $r=0.98$ (Pearson) and $\rho=0.90$ (Spearman)
for chrF++, and at $0.95$ and $0.79$ for BLEU. This is agreement at the level
of whole conditions and says nothing about agreement on individual sentences.

Across the roster, \textsc{inject} raises chrF++ for every one of the ten
models in Komi-Zyrian, Moksha and Erzya, by 1.2 to 8.4 points; under a paired
bootstrap over sentences with 1{,}000 resamples \citep{koehn2004significance}
the gain is significant in 25 of the 30 model-language cells, with no
significant loss. For Udmurt the median change is $-0.6$. Under \textsc{mcp}
the four models that call reliably diverge: out of Komi-Zyrian, Qwen3.8-27B
gains 10.5 points, Llama-3.1-8B loses 0.4, Mistral-Nemo-12B loses 3.2 and
Qwen3-4B loses 11.8. Collapsing the six per-word tools into one sentence-level
call (\textsc{mcp\_coarse}) changes little: 30 model-language cells improve, 29
get worse, and the median change is zero. The thirteen-tool surface
(\textsc{mcp\_rich}, 16 models) raises the reliable callers by 7.0 points on
average, on four cells. The tool-use system prompt alone
(\textsc{mcp\_prompt}) lowers chrF++ by 2.5 points out of Komi-Zyrian and by
1.6 into it.

Into the four languages, every model scores between 2.0 and 5.4 chrF++ against
a floor of 9.3 to 10.8. Lemmatising both sides raises every cell by a few
tenths of a point and raises the floor by as much, which indicates that the
failure in this direction is not a matter of inflection. BLEU and chrF++
differ by an order of magnitude in level, 25.3 chrF++ against 5.7 BLEU out of
Komi-Zyrian, and they order the conditions in the same way in every direction.
Appendix~\ref{sec:permodel} lists every model and direction.

\section{Tool-calling probe}
\label{sec:probe}

\begin{table*}[t]
\centering
\small
\begin{tabular}{l r l l}
\toprule
Model & Size & Tool calling & Evidence \\
\midrule
Qwen3-0.6B & 0.6 & works & probe \\
xLAM-2-1b-fc-r & 1.0 & works & probe \\
gemma-3-1b-it & 1.0 & none & probe, parser search, runs \\
Llama-3.2-1B-Instruct & 1.2 & works & 3899 calls in the runs \\
Hammer2.1-1.5b & 1.5 & unreadable & emits calls no parser reads \\
Qwen2.5-1.5B-Instruct & 1.5 & works & probe \\
Qwen3-1.7B & 1.7 & works & probe \\
EuroLLM-1.7B-Instruct & 1.7 & none & probe, parser search, runs \\
granite-3.3-2b-instruct & 2.0 & works & probe \\
Hammer2.1-3b & 3.0 & unreadable & emits calls no parser reads \\
Qwen2.5-3B-Instruct & 3.0 & works & probe \\
xLAM-2-3b-fc-r & 3.0 & works & probe \\
Llama-3.2-3B-Instruct & 3.2 & works & probe \\
Phi-4-mini-instruct & 3.8 & works & 137 calls in the runs \\
Qwen3-4B & 4.0 & works & probe \\
gemma-3-4b-it & 4.0 & none & probe, parser search, runs \\
Qwen2.5-7B-Instruct & 7.0 & works & probe \\
OLMo-2-1124-7B-Instruct & 7.0 & none & probe, parser search, runs \\
Mistral-7B-Instruct-v0.3 & 7.0 & works & probe \\
c4ai-command-r7b-12-2024 & 8.0 & none & probe, parser search, runs \\
Hermes-3-Llama-3.1-8B & 8.0 & works & probe \\
Qwen3-8B & 8.0 & works & probe \\
granite-3.3-8b-instruct & 8.0 & works & parser llama3\_json \\
Llama-3.1-8B-Instruct & 8.0 & works & probe \\
Ministral-8B-Instruct-2410 & 8.0 & works & probe \\
watt-tool-8B & 8.0 & none & probe, parser search, runs \\
EuroLLM-9B-Instruct & 9.0 & none & probe, parser search, runs \\
gemma-3-12b-it & 12.0 & none & probe, parser search, runs \\
Mistral-Nemo-Instruct-2407 & 12.0 & works & probe \\
gemma-3-27b-it & 27.0 & none & probe, parser search, runs \\
Qwen3.8-27B & 27.8 & works & parser qwen3\_xml \\
\bottomrule
\end{tabular}
\caption{Whether each model can emit a tool call that succeeds. A model counts as capable if the positive control passes, or if an exhaustive parser search finds a parser under which it passes, or if it made successful calls during the sweep -- the control alone gives false negatives, since whether a model calls a tool depends on the prompt.}
\label{tab:capability}
\end{table*}

Table~\ref{tab:capability} records, for each model, whether it can emit a tool
call that succeeds, and on what evidence. A model counts as capable if the
positive control passes, or an exhaustive parser search finds a serving
configuration under which it passes, or it made successful calls during the
sweep. All three are needed: everything was served with vLLM, which wants the
right tool-call parser per model, and the wrong one silently yields zero
parsed calls from a model that is emitting them, while the control alone gives
false negatives because whether a model calls depends on the prompt. Twenty of
the 31 are capable, nine fail all three tests, and two emit call syntax that
no parser in the search reads.

\section{Language codes sent to the tools}
\label{sec:isocodes}

\begin{table*}[t]
\centering
\small
\begin{tabular}{l r r r}
\toprule
Model & Codes & Correct & Unsupported \\
\midrule
Llama-3.1-8B-Instruct & 53513 & 48.4\% & 50.3\% \\
Llama-3.2-3B-Instruct & 40906 & 35.0\% & 36.4\% \\
Qwen3-1.7B & 40682 & 41.9\% & 33.7\% \\
Ministral-8B-Instruct-2410 & 36292 & 38.9\% & 61.1\% \\
Llama-3.2-1B-Instruct & 30530 & 21.8\% & 78.2\% \\
Qwen3-8B & 30207 & 45.1\% & 54.9\% \\
Qwen3.8-27B & 22100 & 74.5\% & 23.5\% \\
Qwen3-4B & 20790 & 24.6\% & 57.1\% \\
Mistral-Nemo-Instruct-2407 & 18079 & 70.7\% & 29.0\% \\
xLAM-2-3b-fc-r & 7642 & 42.6\% & 57.0\% \\
Qwen2.5-3B-Instruct & 6241 & 38.9\% & 60.7\% \\
Qwen2.5-7B-Instruct & 5478 & 20.1\% & 79.8\% \\
granite-3.3-2b-instruct & 1797 & 45.0\% & 55.0\% \\
Qwen2.5-1.5B-Instruct & 1106 & 45.8\% & 54.2\% \\
Hermes-3-Llama-3.1-8B & 304 & 16.8\% & 83.2\% \\
Phi-4-mini-instruct & 156 & 11.5\% & 87.8\% \\
xLAM-2-1b-fc-r & 87 & 11.5\% & 88.5\% \\
\midrule
All & 315910 & 42.5\% & 49.1\% \\
\bottomrule
\end{tabular}
\caption{Language codes the models sent to the tools, counted per call.
\emph{Correct} is a code UralicNLP supports for the language actually being
processed; \emph{unsupported} names nothing the library knows. The two do not
sum to 100: the remainder, 8.4\% over the whole study, is a code the library
supports for the wrong language, which returns an answer about a language
nobody asked for.}
\label{tab:isocodes}
\end{table*}

Table~\ref{tab:isocodes} breaks the figures in Section~\ref{sec:roster} down by
model. The share of codes naming something UralicNLP supports for the text in
hand ranges from 11.5\% to 74.5\%, and no model sends a supported code more
than three times in four. The invalid ones are not random: \texttt{en} is sent
74{,}449 times, and the attempts at Komi-Zyrian run through \texttt{kv},
\texttt{kom}, \texttt{komi-zyrian}, \texttt{kz}, \texttt{kmz} and \texttt{kmm}
before any of them reaches \texttt{kpv}.

\section{Failure modes in full}
\label{sec:failuremodes}

Table~\ref{tab:pathology} classifies every output by what went wrong with it,
rather than by how far it scored from the reference. \emph{Copied} reproduces
the input instead of translating it; \emph{wrong lang.} is decided by script
where the scripts differ and by langdetect where they do not, believing
only the claim that a string is in the source language; \emph{degenerate} loops
or runs past four times the reference length; \emph{empty} returned nothing.
A well-formed output is none of those, which is a test of form and not of
adequacy: it can still be a wrong translation.

\begin{table*}[t]
\centering
\footnotesize
\setlength{\tabcolsep}{4pt}
\begin{tabular}{l l r r r r r}
\toprule
Task & Condition & Well-formed & Copied & Wrong lang. & Degenerate & Empty \\
\midrule
\multirow{7}{*}{Into the endangered languages} & no tools & 72.0 & 3.8 & 11.0 & 13.1 & 0.1 \\
 & tool prompt & 50.1 & 7.8 & 24.0 & 9.5 & 8.5 \\
 & inject & 67.1 & 5.9 & 10.4 & 15.8 & 0.8 \\
 & inject+ & 66.3 & 6.8 & 3.9 & 23.0 & 0.0 \\
 & mcp & 42.5 & 18.5 & 22.9 & 9.2 & 6.8 \\
 & mcp coarse & 43.3 & 11.7 & 25.0 & 11.3 & 8.7 \\
 & mcp rich & 38.9 & 10.4 & 23.0 & 14.3 & 13.5 \\
\midrule
\multirow{7}{*}{Out of the endangered languages} & no tools & 98.1 & 0.3 & 0.1 & 0.6 & 1.0 \\
 & tool prompt & 89.3 & 1.5 & 0.1 & 1.1 & 8.1 \\
 & inject & 93.8 & 0.5 & 0.1 & 0.6 & 5.1 \\
 & inject+ & 96.2 & 0.0 & 0.0 & 0.4 & 3.4 \\
 & mcp & 90.3 & 1.3 & 0.4 & 1.2 & 6.9 \\
 & mcp coarse & 90.8 & 1.1 & 0.1 & 2.2 & 5.8 \\
 & mcp rich & 87.8 & 1.1 & 0.5 & 1.5 & 9.2 \\
\bottomrule
\end{tabular}
\caption{Share of outputs (\%) by failure mode, over every model and sentence
in each group of directions. \emph{Well-formed} is none of the other four, a
test of form and not of adequacy: such an output can still be a wrong
translation. \emph{Copied} reproduces the input, \emph{wrong lang.} is in the
source language, \emph{degenerate} loops or runs past four times the reference
length, \emph{empty} returned nothing. Rows are conditions; the five columns
sum to 100.}
\label{tab:pathology}
\end{table*}

\section{Tokeniser fertility}
\label{sec:fertility}

Table~\ref{tab:fertility} gives subword pieces per word for each tokeniser in
the roster, measured on the parallel set itself, so the six columns describe
the same 68 sentences in six languages.

\begin{table*}[p]
\centering
\scriptsize
\setlength{\tabcolsep}{3pt}
\begin{tabular}{l r r r r r}
\toprule
Tokenizer & eng & kpv & mdf & myv & udm \\
\midrule
Qwen3-0.6B & 1.51 & 3.96 & 3.65 & 3.77 & 3.52 \\
Qwen3-1.7B & 1.51 & 3.96 & 3.65 & 3.77 & 3.52 \\
Qwen3-4B & 1.51 & 3.96 & 3.65 & 3.77 & 3.52 \\
Qwen3-8B & 1.51 & 3.96 & 3.65 & 3.77 & 3.52 \\
Llama-3.2-1B-Instruct & 1.51 & 4.24 & 3.44 & 3.62 & 3.48 \\
Llama-3.2-3B-Instruct & 1.51 & 4.24 & 3.44 & 3.62 & 3.48 \\
Llama-3.1-8B-Instruct & 1.51 & 4.24 & 3.44 & 3.62 & 3.48 \\
gemma-3-1b-it & 1.42 & 3.50 & 3.21 & 3.27 & 3.10 \\
gemma-3-4b-it & 1.42 & 3.50 & 3.21 & 3.27 & 3.10 \\
gemma-3-12b-it & 1.42 & 3.50 & 3.21 & 3.27 & 3.10 \\
Ministral-8B-Instruct-2410 & 1.58 & 4.13 & 3.33 & 3.37 & 3.33 \\
Phi-4-mini-instruct & 1.48 & 4.01 & 3.24 & 3.39 & 3.21 \\
OLMo-2-1124-7B-Instruct & 1.52 & 4.98 & 4.41 & 4.53 & 4.17 \\
EuroLLM-1.7B-Instruct & 1.36 & 4.09 & 3.16 & 3.37 & 3.36 \\
EuroLLM-9B-Instruct & 1.59 & 4.13 & 3.22 & 3.34 & 3.38 \\
Viking-7B & 1.55 & 4.67 & 4.26 & 4.34 & 3.91 \\
\bottomrule
\end{tabular}
\caption{Subword pieces per word, measured on the parallel set itself, so the
six columns describe the same 68 sentences in six languages. A value of 1.0
would mean one token per word; English costs 1.36 to 1.59 and the endangered languages 2.0 to 3.3 times
that.}
\label{tab:fertility}
\end{table*}

\section{Extraction}
\label{sec:extraction}

Of 62{,}920 records, 40{,}783 raw outputs were already a single clean line
under 400 characters with no markup and were scored unchanged; 4{,}072 were
empty; the remaining 18{,}065 were read by a language model instructed to
return the translation inside each one as a verbatim substring, stripping
code fences, XML tags, JSON envelopes, lead-ins, trailing commentary, role
markers, emphasis and enclosing quotes, while keeping dialogue dashes,
in-sentence quotes and degenerate repetition. All 18{,}065 returned values
passed verbatim containment against their source, so the extractor removed
packaging and never wrote text of its own. Replacing all 4{,}541 empty strings
with the whole raw output moves the headline gap by at most 0.5 chrF++, so
nothing in the paper turns on how an answerless output is treated. The full
instruction, the inputs, the returned values and the validation code are
released with the data.

\section{Every model, every direction}
\label{sec:permodel}

These tables list every cell behind the averages in Appendix~\ref{sec:scores}. Every cell the paper averages over is listed
here: ten models, five directions each way, under the conditions that carry
the argument.

\begin{table*}[p]
\centering\scriptsize
\setlength{\tabcolsep}{4pt}
\setlength{\aboverulesep}{1pt}\setlength{\belowrulesep}{1pt}
\begin{tabular}{@{}l rrr rrr@{}}
\toprule
Model & zs & inj & mcp & zs & inj & mcp \\
\midrule
\multicolumn{7}{@{}l}{\itshape kpv--eng}\\
granite-3.3-8b-instruct & 18.8 & 22.7 & 17.1 & 1.6 & 1.9 & 1.6 \\
Hermes-3-Llama-3.1-8B & 23.0 & 27.8 & 22.5 & 5.2 & 6.2 & 4.9 \\
Llama-3.1-8B-Instruct & 21.7 & 29.5 & 21.2 & 3.7 & 9.5 & 3.8 \\
Mistral-7B-Instruct-v0.3 & 17.3 & 24.8 & 17.7 & 1.8 & 3.7 & 2.2 \\
Mistral-Nemo-Instruct-2407 & 16.9 & 22.5 & 13.7 & 2.2 & 3.5 & 1.6 \\
Qwen2.5-7B-Instruct & 19.2 & 27.9 & 17.9 & 3.4 & 6.9 & 2.6 \\
Qwen3-0.6B & 15.2 & 14.5 & 10.8 & 1.4 & 1.1 & 0.5 \\
Qwen3-4B & 17.4 & 19.9 & 5.7 & 1.6 & 2.1 & 0.1 \\
Qwen3-8B & 18.9 & 25.9 & 17.8 & 4.0 & 5.2 & 2.9 \\
Qwen3.8-27B & 29.9 & 37.8 & 40.5 & 10.2 & 16.3 & 19.6 \\
\addlinespace[2pt]
\multicolumn{7}{@{}l}{\itshape mdf--eng}\\
granite-3.3-8b-instruct & 17.8 & 21.4 & 16.8 & 2.1 & 2.8 & 1.8 \\
Hermes-3-Llama-3.1-8B & 19.2 & 24.0 & 17.4 & 1.9 & 3.9 & 2.0 \\
Llama-3.1-8B-Instruct & 19.5 & 22.9 & 17.0 & 2.7 & 3.5 & 2.7 \\
Mistral-7B-Instruct-v0.3 & 17.6 & 21.4 & 16.7 & 1.4 & 2.1 & 1.6 \\
Mistral-Nemo-Instruct-2407 & 17.7 & 22.7 & 13.2 & 1.4 & 3.2 & 1.1 \\
Qwen2.5-7B-Instruct & 17.6 & 22.8 & 18.9 & 2.0 & 3.2 & 3.3 \\
Qwen3-0.6B & 5.9 & 13.3 & 4.5 & 0.2 & 2.0 & 0.1 \\
Qwen3-4B & 18.6 & 19.0 & 9.3 & 3.3 & 3.5 & 0.2 \\
Qwen3-8B & 18.1 & 23.2 & 6.7 & 3.2 & 4.3 & 0.3 \\
Qwen3.8-27B & 22.4 & 30.2 & 31.6 & 4.9 & 7.8 & 11.2 \\
\addlinespace[2pt]
\multicolumn{7}{@{}l}{\itshape myv--eng}\\
granite-3.3-8b-instruct & 16.0 & 19.9 & 15.4 & 1.2 & 1.3 & 1.1 \\
Hermes-3-Llama-3.1-8B & 10.6 & 21.8 & 17.1 & 0.3 & 2.8 & 1.2 \\
Llama-3.1-8B-Instruct & 17.9 & 21.9 & 15.7 & 1.9 & 3.0 & 2.2 \\
Mistral-7B-Instruct-v0.3 & 14.7 & 20.1 & 13.9 & 1.1 & 1.4 & 1.3 \\
Mistral-Nemo-Instruct-2407 & 14.8 & 19.2 & 13.7 & 1.4 & 1.6 & 0.7 \\
Qwen2.5-7B-Instruct & 15.8 & 20.6 & 15.5 & 1.7 & 1.9 & 1.5 \\
Qwen3-0.6B & 9.3 & 9.1 & 4.2 & 0.2 & 0.3 & 0.1 \\
Qwen3-4B & 16.0 & 16.7 & 8.2 & 1.5 & 3.7 & 0.5 \\
Qwen3-8B & 15.9 & 21.3 & 10.5 & 1.5 & 3.1 & 0.3 \\
Qwen3.8-27B & 20.2 & 29.7 & 30.9 & 3.8 & 8.7 & 9.7 \\
\addlinespace[2pt]
\multicolumn{7}{@{}l}{\itshape udm--eng}\\
granite-3.3-8b-instruct & 16.6 & 16.9 & 15.7 & 1.7 & 1.7 & 1.3 \\
Hermes-3-Llama-3.1-8B & 19.0 & 17.3 & 18.3 & 3.3 & 2.3 & 1.4 \\
Llama-3.1-8B-Instruct & 18.4 & 16.8 & 14.6 & 3.6 & 2.4 & 1.6 \\
Mistral-7B-Instruct-v0.3 & 15.5 & 14.4 & 15.6 & 1.4 & 1.0 & 1.9 \\
Mistral-Nemo-Instruct-2407 & 15.3 & 14.8 & 12.0 & 1.5 & 1.4 & 0.3 \\
Qwen2.5-7B-Instruct & 15.3 & 15.4 & 15.3 & 2.1 & 1.6 & 1.9 \\
Qwen3-0.6B & 6.0 & 12.7 & 5.8 & 0.2 & 1.3 & 0.2 \\
Qwen3-4B & 15.8 & 15.2 & 5.3 & 1.9 & 1.7 & 0.5 \\
Qwen3-8B & 16.2 & 14.8 & 15.8 & 2.5 & 1.4 & 2.0 \\
Qwen3.8-27B & 24.5 & 26.2 & 29.7 & 7.3 & 7.2 & 10.2 \\
\bottomrule
\end{tabular}
\caption{Every model and every direction into English, scored on the model extraction. The first three columns are chrF++ and the last three BLEU. The chrF++ floor is 11.5.}
\label{tab:permodel-out}
\end{table*}

\begin{table*}[p]
\centering\scriptsize
\setlength{\tabcolsep}{4pt}
\setlength{\aboverulesep}{1pt}\setlength{\belowrulesep}{1pt}
\begin{tabular}{@{}l rrrr rrrr@{}}
\toprule
Model & zs & inj & mcp & crs & zs & inj & mcp & crs \\
\midrule
\multicolumn{9}{@{}l}{\itshape eng--kpv}\\
granite-3.3-8b-instruct & 5.9 & 3.3 & 1.8 & 2.2 & 6.2 & 3.4 & 1.9 & 2.3 \\
Hermes-3-Llama-3.1-8B & 7.0 & 7.8 & 6.6 & 5.7 & 7.2 & 8.2 & 6.8 & 6.2 \\
Llama-3.1-8B-Instruct & 2.4 & 2.1 & 2.9 & 2.0 & 2.4 & 2.3 & 3.2 & 2.1 \\
Mistral-7B-Instruct-v0.3 & 3.0 & 3.6 & 2.9 & 1.5 & 3.2 & 3.5 & 3.1 & 1.6 \\
Mistral-Nemo-Instruct-2407 & 3.9 & 4.3 & 1.8 & 2.2 & 4.0 & 4.5 & 2.0 & 2.7 \\
Qwen2.5-7B-Instruct & 8.6 & 3.9 & 3.8 & 5.1 & 8.8 & 3.9 & 4.1 & 5.3 \\
Qwen3-0.6B & 0.4 & 1.8 & 0.5 & 0.1 & 0.4 & 2.2 & 0.5 & 0.1 \\
Qwen3-4B & 1.4 & 1.8 & 1.3 & 1.9 & 1.5 & 1.8 & 1.4 & 2.0 \\
Qwen3-8B & 2.4 & 1.7 & 0.5 & 0.3 & 2.4 & 1.7 & 0.6 & 0.3 \\
Qwen3.8-27B & 9.0 & 6.8 & 22.6 & 5.0 & 9.2 & 6.9 & 26.3 & 5.2 \\
\addlinespace[2pt]
\multicolumn{9}{@{}l}{\itshape eng--mdf}\\
granite-3.3-8b-instruct & 1.6 & 0.8 & 1.2 & 0.8 & 1.6 & 0.8 & 1.2 & 0.8 \\
Hermes-3-Llama-3.1-8B & 3.1 & 4.5 & 3.5 & 2.4 & 3.1 & 4.6 & 3.3 & 2.4 \\
Llama-3.1-8B-Instruct & 0.4 & 0.5 & 2.1 & 1.7 & 0.4 & 0.5 & 2.1 & 1.6 \\
Mistral-7B-Instruct-v0.3 & 1.3 & 0.7 & 2.5 & 1.4 & 1.3 & 0.8 & 2.6 & 1.4 \\
Mistral-Nemo-Instruct-2407 & 3.7 & 1.9 & 1.4 & 1.9 & 3.9 & 2.1 & 1.5 & 1.9 \\
Qwen2.5-7B-Instruct & 9.6 & 4.1 & 3.2 & 2.3 & 10.1 & 4.2 & 3.3 & 2.4 \\
Qwen3-0.6B & 1.4 & 1.6 & 2.5 & 2.5 & 1.4 & 1.6 & 2.6 & 2.6 \\
Qwen3-4B & 1.4 & 1.2 & 1.1 & 0.6 & 1.4 & 1.3 & 1.2 & 0.6 \\
Qwen3-8B & 1.8 & 0.7 & 0.5 & 0.4 & 1.9 & 0.7 & 0.5 & 0.4 \\
Qwen3.8-27B & 10.2 & 4.8 & 17.6 & 5.8 & 10.7 & 5.3 & 20.7 & 5.9 \\
\addlinespace[2pt]
\multicolumn{9}{@{}l}{\itshape eng--myv}\\
granite-3.3-8b-instruct & 1.6 & 1.6 & 1.4 & 1.7 & 1.5 & 1.5 & 1.4 & 1.6 \\
Hermes-3-Llama-3.1-8B & 5.8 & 5.3 & 3.0 & 3.3 & 5.7 & 5.3 & 3.1 & 3.2 \\
Llama-3.1-8B-Instruct & 1.7 & 1.6 & 5.0 & 2.0 & 1.6 & 1.6 & 4.8 & 2.0 \\
Mistral-7B-Instruct-v0.3 & 2.4 & 2.8 & 1.8 & 0.8 & 2.4 & 2.9 & 1.7 & 0.7 \\
Mistral-Nemo-Instruct-2407 & 5.5 & 2.5 & 2.5 & 1.4 & 5.6 & 2.4 & 2.7 & 1.5 \\
Qwen2.5-7B-Instruct & 9.1 & 3.7 & 6.5 & 8.5 & 9.5 & 3.6 & 7.7 & 8.8 \\
Qwen3-0.6B & 1.1 & 1.2 & 0.9 & 0.7 & 1.2 & 1.3 & 1.0 & 0.7 \\
Qwen3-4B & 1.9 & 2.8 & 1.1 & 2.0 & 1.9 & 2.9 & 1.1 & 2.0 \\
Qwen3-8B & 1.2 & 1.3 & 0.5 & 0.3 & 1.2 & 1.4 & 0.5 & 0.4 \\
Qwen3.8-27B & 9.3 & 6.4 & 21.7 & 6.5 & 9.5 & 6.7 & 26.0 & 6.8 \\
\addlinespace[2pt]
\multicolumn{9}{@{}l}{\itshape eng--udm}\\
granite-3.3-8b-instruct & 3.6 & 3.1 & 1.1 & 2.8 & 3.8 & 3.1 & 1.2 & 2.9 \\
Hermes-3-Llama-3.1-8B & 5.8 & 4.0 & 5.0 & 4.7 & 6.2 & 4.2 & 5.5 & 5.0 \\
Llama-3.1-8B-Instruct & 3.4 & 2.0 & 3.4 & 3.0 & 3.5 & 2.1 & 3.6 & 3.3 \\
Mistral-7B-Instruct-v0.3 & 2.5 & 3.9 & 2.1 & 1.1 & 2.5 & 4.1 & 2.1 & 1.1 \\
Mistral-Nemo-Instruct-2407 & 5.2 & 3.0 & 3.3 & 3.0 & 5.7 & 3.2 & 3.4 & 3.1 \\
Qwen2.5-7B-Instruct & 11.6 & 7.6 & 5.8 & 8.8 & 12.3 & 8.2 & 5.9 & 9.4 \\
Qwen3-0.6B & 0.1 & 0.1 & 0.4 & 0.3 & 0.1 & 0.1 & 0.4 & 0.3 \\
Qwen3-4B & 8.0 & 4.4 & 1.3 & 4.6 & 8.1 & 4.5 & 1.4 & 4.7 \\
Qwen3-8B & 1.5 & 1.3 & 0.7 & 0.7 & 1.6 & 1.4 & 0.7 & 0.7 \\
Qwen3.8-27B & 12.0 & 10.8 & 7.8 & 7.7 & 12.7 & 12.0 & 8.2 & 7.9 \\
\bottomrule
\end{tabular}
\caption{Every model and every direction into a Uralic language, scored on the model extraction. The first four columns are chrF++ and the last four the same measure on lemmatised text. Floors are 9.3 to 10.8 chrF++ and 9.9 to 11.5 lemmatised.}
\label{tab:permodel-into}
\end{table*}

\end{document}